\documentclass[runningheads]{llncs}
\usepackage[T1]{fontenc}
\usepackage{hyperref}
\usepackage{multirow}
\usepackage{xcolor}
\definecolor{eccvblue}{rgb}{0.12,0.49,0.85}
\hypersetup{colorlinks=true,citecolor=eccvblue}
\usepackage{graphicx}
\usepackage{amssymb}

\begin{document}
\title{SAM2-UNet: Segment Anything 2 Makes Strong Encoder for Natural and Medical Image Segmentation}
\titlerunning{SAM2-UNet}
\author{Xinyu Xiong\inst{1*} \and
Zihuang Wu\inst{2*} \and Shuangyi Tan\inst{3} \and Wenxue Li\inst{4} \and \\ Feilong Tang\inst{5} \and Ying Chen\inst{6}, Siying Li\inst{7} \and Jie Ma\inst{1} \and Guanbin Li\inst{1\dagger}
}

\authorrunning{X. Xiong et al.}
\institute{$^1$School of Computer Science and Engineering, Sun Yat-sen University \\
$^2$School of Computer and Information Engineering, Jiangxi Normal University \\
$^3$The Chinese University of Hong Kong (Shenzhen) \\
$^4$Tianjin University $^5$Monash University $^6$Pazhou Lab \\
$^7$Smart Hospital Research Institute, Peking University Shenzhen Hospital
}

\renewcommand{\thefootnote}{}
\footnotetext{\inst{*} Authors contributed equally to this work.}
\footnotetext{\inst{\dagger} Corresponding author.}
\maketitle 

\begin{abstract}
Image segmentation plays an important role in vision understanding. Recently, the emerging vision foundation models continuously achieved superior performance on various tasks. Following such success, in this paper, we prove that the Segment Anything Model 2 (SAM2) can be a strong encoder for U-shaped segmentation models. We propose a simple but effective framework, termed SAM2-UNet, for versatile image segmentation. Specifically, SAM2-UNet adopts the Hiera backbone of SAM2 as the encoder, while the decoder uses the classic U-shaped design. Additionally, adapters are inserted into the encoder to allow parameter-efficient fine-tuning. Preliminary experiments on various downstream tasks, such as camouflaged object detection, salient object detection, marine animal segmentation, mirror detection, and polyp segmentation, demonstrate that our SAM2-UNet can simply beat existing specialized state-of-the-art methods without bells and whistles. Project page: \url{https://github.com/WZH0120/SAM2-UNet}.
\end{abstract}

\section{Introduction}

Image segmentation is a crucial task in the field of computer vision, serving as the foundation for various visual understanding applications. By dividing an image into meaningful regions based on specific semantic criteria, image segmentation enables a wide array of downstream tasks in both natural and medical domains, such as camouflaged object detection~\cite{IJCAI21_C2FNet,CVPR22_ZoomNet}, salient object detection~\cite{AAAI20_F3Net,CVPR23_FEDER}, marine animal segmentation~\cite{JOE23_MASNet,TCSVT21_MAS3K}, mirror detection~\cite{AAAI23_HetNet,CVPR22_SAMirror}, and polyp segmentation~\cite{MICCAI20_PraNet,PR23_CFANet}. Many specialized architectures have been proposed to achieve superior performance on these different tasks, while it remains an open challenge to design a unified architecture to address the diverse segmentation tasks.

The emergence of vision foundation models (VFMs)~\cite{ICCV23_SAM1,SAM2,ICCV23_SegGPT,CVPR24_OMGSeg} has introduced significant potential in the field of image segmentation. Among these VFMs, a notable example is the Segment Anything Model (SAM1)~\cite{ICCV23_SAM1} and its successor, Segment Anything 2 (SAM2)~\cite{SAM2}. SAM2 builds upon the foundation laid by SAM1, utilizing a larger dataset for training and incorporating improvements in architectural design. However, despite these advancements, SAM2 still produces class-agnostic segmentation results when no manual prompt is provided. This limitation highlights the ongoing challenge of effectively transferring SAM2 to downstream tasks, where task-specific or class-specific segmentation is often required. Exploring strategies to enhance SAM2's adaptability and performance in these scenarios remains an important area of research.

To adapt SAM to downstream tasks, several approaches have been proposed, including the use of adapters~\cite{ICCVW23_SAMAdapter,arXiv23_SAMed} for parameter-efficient fine-tuning and the integration of additional conditional inputs such as text prompts~\cite{CVPR24_AlignSAM,arXiv24_EVFSAM,MICCAI24_TPDRSeg} or in-context samples~\cite{ICLR24_PerSAM,ICLR24_Matcher}. Inspired by the strong segmentation capabilities of U-Net~\cite{MICCAI15_UNet} and its variants~\cite{TMI19_UNetPP,MedIA24_TransUNet,ECCVW22_SwinUNet}, some researchers have explored the possibility of transforming SAM into a U-shaped architecture~\cite{MICCAI24_DeSAM,MLMI23_MammoSAM}. However, these efforts have often been limited by the plain structure of the vanilla ViT encoder~\cite{ICLR21_ViT}, which lacks the hierarchy needed for more sophisticated segmentation tasks. Fortunately, the introduction of SAM2, which features a hierarchical backbone, opens new avenues for designing a U-shaped network with improved effectiveness.

In this paper, we propose SAM2-UNet, the benefit of which is summarized as follows:
\begin{itemize}
    \item \textbf{Simplicity.} SAM2-UNet adopts a classic U-shaped encoder-decoder architecture, known for its ease of use and high extensibility.

    \item \textbf{Efficiency.} Adapters are integrated into the encoder to enable parameter-efficient fine-tuning, allowing the model to be trained even on memory-limited devices.
    
    \item \textbf{Effectiveness.} Extensive experiments on eighteen public datasets demonstrate that SAM2-UNet delivers powerful performance across five challenging benchmarks.
\end{itemize}

\section{Method}
\begin{figure*}[t]
    \centering
    \includegraphics[width=1.0\linewidth]{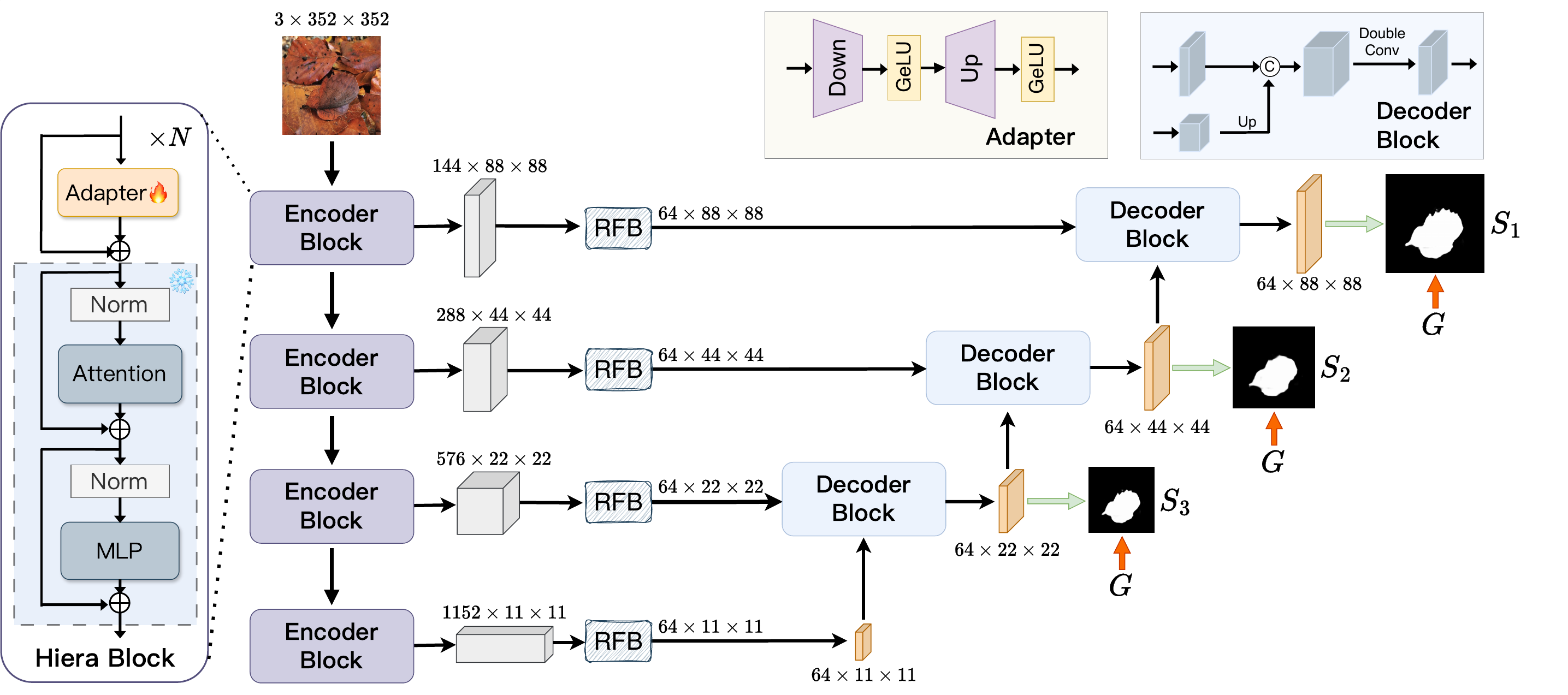}
    \caption{Overview of the proposed SAM2-UNet. Note that there are some variants of the Hiera block, and we only demonstrate a simplified structure for ease of understanding.} 
    \label{fig:sam2unet}
\end{figure*}

The overall architecture of SAM2-UNet is illustrated in Fig.~\ref{fig:sam2unet}, comprising four main components: encoder, decoder, receptive field blocks (RFBs), and adapters. Note that we discard components that are not essential for constructing a basic U-Net~\cite{MICCAI15_UNet}, such as memory attention, prompt encoder, memory encoder, and memory bank.

\textbf{Encoder.} SAM2-UNet applys the Hiera~\cite{ICML23_Hiera} backbone pretrained by SAM2. Compared with the plain ViT~\cite{ICLR21_ViT} encoder used in SAM1~\cite{ICCV23_SAM1}, Hiera uses a hierarchical structure that allows multiscale feature capturing, which is more suitable for designing a U-shaped network. Specifically, given an input image $I \in \mathbb{R}^{3 \times H \times W}$, where $H$ denotes height and $W$ denotes width, Hiera will output four hierarchical features $X_i \in \mathbb{R}^{C_i \times \frac{H}{2^{i+1}} \times \frac{W}{2^{i+1}}} (i \in \{1, 2, 3, 4\})$. For Hiera-L, $C_i \in \{144, 288, 576, 1152\}$.

\textbf{RFBs.} After extracting the encoder features, we pass them through four receptive field blocks~\cite{ECCV18_RFB,MICCAI20_PraNet} to reduce the channel number to 64 as well as enhance these lightweight features.

\textbf{Adapters.} As the parameters of Hiera may be huge (214M for Hiera-L), performing full fine-tuning would not always be memory feasible. Therefore, we freeze the parameters of Hiera and insert adapters before each multi-scale block of Hiera to achieve parameter-efficient fine-tuning. Similar to the adapter design in~\cite{ICML19_Adapter,arXiv23_Learnable}, each adapter in our framework consists of a linear layer for downsampling, a GeLU activation function, followed by another linear layer for upsampling, and a final GeLU activation.

\textbf{Decoder.} The original mask decoder in SAM2 uses a two-way transformer approach to facilitate feature interaction between the prompt embedding and encoder features. In contrast, inspired by the highly customizable U-shaped structure that has proven effective in many tasks~\cite{TMI19_UNetPP,MedIA24_TransUNet,ECCVW22_SwinUNet}, our decoder also adheres to the classic U-Net design. It consists of three decoder blocks, each containing two `Conv-BN-ReLU' combinations, where `Conv' denotes a $3 \times 3$ convolution layer and `BN' represents batch normalization. The output feature from each decoder block passes through a $1 \times 1$ Conv segmentation head to produce a segmentation result $S_i$ ($i \in {1, 2, 3}$), which is then upsampled and supervised by the ground truth mask $G$.

\textbf{Loss Function.} Following the approaches in~\cite{MICCAI20_PraNet,AAAI20_F3Net}, we use the weighted IoU loss and binary cross-entropy (BCE) loss as our training objectives: $\mathcal{L} = \mathcal {L}_{IoU}^w + \mathcal {L}_{BCE}^w$. Additionally, we apply deep supervision to all segmentation outputs $S_i$. The total loss for SAM2-UNet is formulated as: $\mathcal{L}_{total} = \sum_{i=1}^{3} \mathcal{L}(G, S_i)$.

\begin{table}[t]
\centering
\caption{Detailed information of datasets for different tasks.}
\label{tab:dataset}
\renewcommand\arraystretch{1.2}
\renewcommand\tabcolsep{2pt}
\begin{tabular}{l|c|cc}
\hline
    Tasks & Dataset  & Train Set & Test Set \\
        \hline
    & CAMO~\cite{CAMO} & 1,000 & 250 \\
    & COD10K~\cite{CVPR20_SINet} & 3,040 & 2,026 \\
    & CHAMELEON~\cite{chameleon} & - & 76 \\
    \multirow{-4}{*}{Camouflaged Object Detection} & NC4K~\cite{CVPR21_NC4K} & - &  4,121 \\
    \hline
    & DUTS~\cite{CVPR17_DUTS} & 10,553 & 5,019 \\
    & DUT-OMRON~\cite{CVPR13_DUTO} & - & 5,168\\
    & HKU-IS~\cite{CVPR15_HKUIS} & - & 4,447\\
    & PASCAL-S~\cite{CVPR14_PASCALS} & - & 850 \\
    \multirow{-5}{*}{Salient Object Detection} & 
    ECSSD~\cite{CVPR13_ECSSD} & - & 1,000\\
    \hline
     & MAS3K~\cite{TCSVT21_MAS3K} & 1,769 & 1,141 \\
    \multirow{-2}{*}{Marine Animal Segmentation} & RMAS~\cite{JOE23_MASNet} & 2,514 & 500 \\
    \hline
     & MSD~\cite{ICCV19_MirrorNet} & 3,063 & 955 \\
    \multirow{-2}{*}{Mirror Detection} & PMD~\cite{CVPR20_PMD} & 5,096 & 571 \\
    \hline
    & Kvasir-SEG~\cite{MMM20_KvasirSEG} & 900 & 100 \\
    & CVC-ClinicDB~\cite{CMIG15_ClinicDB} & 550 & 62 \\
    & CVC-ColonDB~\cite{TMI15_ColonDB} & - & 380 \\
    & CVC-300~\cite{Endoscene300} & - & 60 \\
    \multirow{-5}{*}{Polyp Segmentation} & ETIS~\cite{ETIS} & - & 196  \\
    \hline
\end{tabular}
\end{table}

\section{Experiments}
\begin{figure*}[t]
    \centering
    \includegraphics[width=0.98\linewidth]{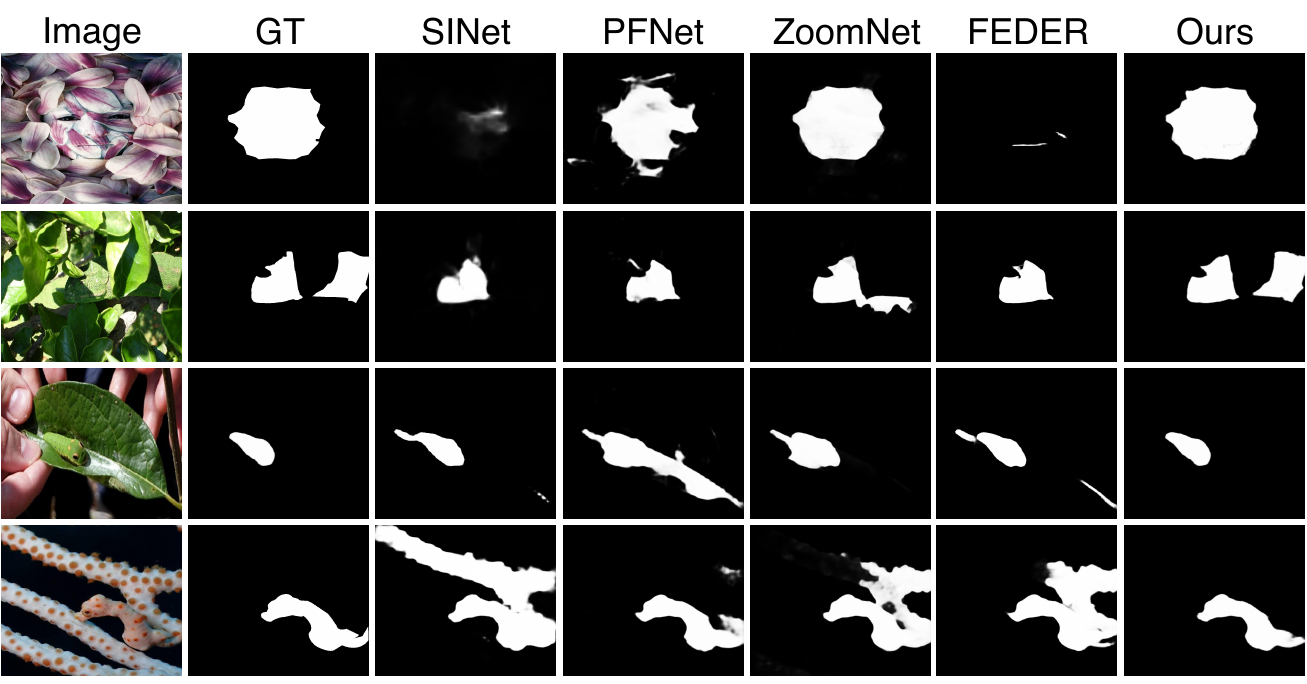}
    \caption{Visualization results on camouflaged object detection.} 
    \label{fig:cod}
\end{figure*}
\begin{figure*}[t]
    \centering
    \includegraphics[width=0.98\linewidth]{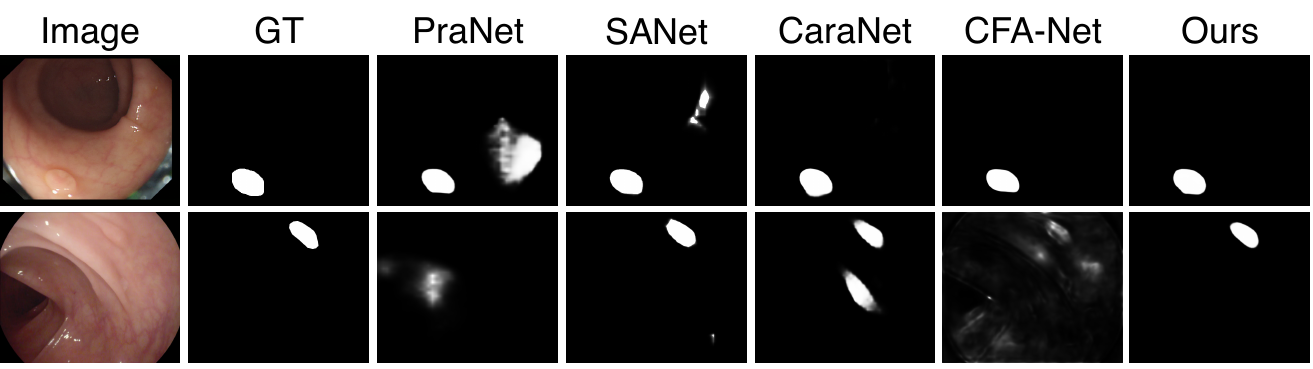}
    \caption{Visualization results on polyp segmentation.} 
    \label{fig:polyp}
\end{figure*}
\subsection{Datasets and Benchmarks}
Our experiments are conducted on five different benchmarks with eighteen datasets in total, as shown in Table~\ref{tab:dataset}:

\textbf{Camouflaged Object Detection} aims to detect objects well hidden in the environment. We adopt four datasets for benchmarking, including CAMO~\cite{CAMO}, COD10K~\cite{CVPR20_SINet}, CHAMELEON~\cite{chameleon}, and NC4K~\cite{CVPR21_NC4K}. Four metrics are used for comparison, including S-measure ($S_\alpha$)~\cite{CVPR17_Smeasure}, adaptive F-measure ($F_{\beta}$)~\cite{CVPR14_Fmeasure}, mean E-measure ($E_{\phi}$)~\cite{Emeasure}, and mean absolute error (MAE).

\textbf{Salient Object Detection} aims to mimic human cognition mechanisms to identify salient objects. We adopt five datasets for benchmarking, including DUTS~\cite{CVPR17_DUTS}, DUT-O~\cite{CVPR13_DUTO}, HKU-IS~\cite{CVPR15_HKUIS}, PASCAL-S~\cite{CVPR14_PASCALS}, and ECSSD~\cite{CVPR13_ECSSD}. Three metrics are used for comparison, including S-measure ($S_\alpha$)~\cite{CVPR17_Smeasure}, mean E-measure ($E_{\phi}$)~\cite{Emeasure}, and mean absolute error (MAE).

\textbf{Marine Animal Segmentation} focuses on exploring underwater environments to find marine animals. We adopt two datasets for benchmarking, including MAS3K~\cite{TCSVT21_MAS3K} and RMAS~\cite{JOE23_MASNet}. Five metrics are used for comparison, including mIoU, S-measure ($S_\alpha$)~\cite{CVPR17_Smeasure}, weighted F-measure ($F_{\beta}^{w}$)~\cite{CVPR14_Fmeasure}, mean E-measure ($E_{\phi}$)~\cite{Emeasure}, and mean absolute error (MAE).

\textbf{Mirror Detection} can identify the mirror regions in the given input image. We adopt two datasets for benchmarking, including MSD~\cite{ICCV19_MirrorNet} and PMD~\cite{CVPR20_PMD}. Three metrics are used for comparison, including IoU, F-measure~\cite{CVPR14_Fmeasure}, and mean absolute error (MAE).

\textbf{Polyp Segmentation} helps in the diagnosis of colorectal cancer. We adopt five datasets for benchmarking, including Kvasir-SEG~\cite{MMM20_KvasirSEG}, CVC-ClincDB~\cite{CMIG15_ClinicDB}, CVC-ColonDB~\cite{TMI15_ColonDB}, CVC-300~\cite{Endoscene300}, and ETIS~\cite{ETIS}. Two metrics are used for comparison, including mean Dice (mDice) and mean IoU (mIoU).

\begin{table}[!htpb]
\centering
\caption{Camouflaged object detection performance on CHAMELEON~\cite{chameleon} and CAMO~\cite{CAMO} datasets.}
\label{tab:cod1}
\renewcommand\arraystretch{1.2}
\renewcommand\tabcolsep{2pt}
\begin{tabular}{l|cccc|cccc}
\hline
    & 
    \multicolumn{4}{c|}{CHAMELEON} & \multicolumn{4}{c}{CAMO} \\
    \multirow{-2}{*}{Methods} & $S_\alpha$ & $F_{\beta}$ & $E_{\phi}$ & MAE & $S_\alpha$ & $F_{\beta}$ & $E_{\phi}$ & MAE \\
    \hline
    SINet~\cite{CVPR20_SINet} & 0.872 & 0.823 & 0.936 & 0.034 & 0.745 & 0.712 & 0.804 & 0.092\\
    PFNet~\cite{CVPR21_PFNet} & 0.882 & 0.820 & 0.931 & 0.033 & 0.782 & 0.751 & 0.841 & 0.085 \\
    ZoomNet~\cite{CVPR22_ZoomNet} & 0.902 & 0.858 & 0.943 & 0.024 & 0.820 & 0.792 & 0.877 & 0.066 \\
    FEDER~\cite{CVPR23_FEDER} & 0.903 & 0.856 & 0.947 & 0.026 & 0.836 & 0.807 & 0.897 & 0.066 \\
    \hline
    \textbf{SAM2-UNet} & \textbf{0.914} & \textbf{0.863} & \textbf{0.961} & \textbf{0.022} & \textbf{0.884} & \textbf{0.861} & \textbf{0.932} & \textbf{0.042} \\  
    \hline
\end{tabular}
\end{table}

\begin{table}[!htpb]
\centering
\caption{Camouflaged object detection performance on COD10K~\cite{CVPR20_SINet} and NC4K~\cite{CVPR21_NC4K} datasets.}
\label{tab:cod2}
\renewcommand\arraystretch{1.2}
\renewcommand\tabcolsep{2pt}
\begin{tabular}{l|cccc|cccc}
\hline
    & 
    \multicolumn{4}{c|}{COD10K} & \multicolumn{4}{c}{NC4K} \\
    \multirow{-2}{*}{Methods} & $S_\alpha$ & $F_{\beta}$ & $E_{\phi}$ & MAE & $S_\alpha$ & $F_{\beta}$ & $E_{\phi}$ & MAE \\
    \hline
    SINet~\cite{CVPR20_SINet} & 0.776 & 0.667 & 0.864 & 0.043 & 0.808 & 0.768 & 0.871 & 0.058\\
    PFNet~\cite{CVPR21_PFNet} & 0.800 & 0.676 & 0.877 & 0.040 & 0.829 & 0.779 & 0.887 & 0.053 \\
    ZoomNet~\cite{CVPR22_ZoomNet} & 0.838 & 0.740 & 0.888 & 0.029 &  0.853 & 0.814 & 0.896 & 0.043\\
    FEDER~\cite{CVPR23_FEDER} & 0.844 & 0.748 &  0.911 & 0.029 & 0.862 &  0.824 &  0.913 &  0.042\\
    \hline
    \textbf{SAM2-UNet} & \textbf{0.880} & \textbf{0.789} & \textbf{0.936}  & \textbf{0.021} & \textbf{0.901}  & \textbf{0.863} & \textbf{0.941} & \textbf{0.029}\\  
    \hline
\end{tabular}
\end{table}

\begin{table}[!htpb]
\centering
\caption{Salient object detection performance on DUTS-TE~\cite{CVPR17_DUTS}, DUT-OMRON~\cite{CVPR13_DUTO}, and HKU-IS~\cite{CVPR15_HKUIS} datasets.}
\label{tab:saliency1}
\renewcommand\arraystretch{1.2}
\renewcommand\tabcolsep{2pt}
\begin{tabular}{l|ccc|ccc|ccc}
\hline
    & 
    \multicolumn{3}{c|}{DUTS-TE} & \multicolumn{3}{c|}{DUT-OMRON} & \multicolumn{3}{c}{HKU-IS}\\
    \multirow{-2}{*}{Methods} & $S_\alpha$ & $E_{\phi}$ & MAE & $S_\alpha$ & $E_{\phi}$ & MAE & $S_\alpha$ & $E_{\phi}$ & MAE \\
    \hline
    U2Net~\cite{PR20_U2Net} & 0.874  & 0.884 & 0.044 & 0.847  & 0.872 & 0.054 & 0.916  & 0.948 & 0.031 \\
    ICON~\cite{TPAMI22_ICON} & 0.889  & 0.914 & 0.037 & 0.845  & 0.879 & 0.057 & 0.920  & 0.959 & 0.029\\
    EDN~\cite{TIP22_EDN} & 0.892  & 0.925 & 0.035 & 0.850 & 0.877  & 0.049 & 0.924 & 0.955 & 0.026 \\
    MENet~\cite{CVPR23_MENet} & 0.905  & 0.937 & 0.028 & 0.850  & 0.891 & 0.045 & 0.927 & 0.966 & 0.023 \\
    \hline
    \textbf{SAM2-UNet} & \textbf{0.934} & \textbf{0.959} & \textbf{0.020} & \textbf{0.884} & \textbf{0.912} & \textbf{0.039} & \textbf{0.941} & \textbf{0.971} & \textbf{0.019} \\
    \hline
\end{tabular}
\end{table}

\begin{table}[!htpb]
\centering
\caption{Salient object detection performance on PASCAL-S~\cite{CVPR14_PASCALS} and ECSSD~\cite{CVPR13_ECSSD} datasets.}
\label{tab:saliency2}
\renewcommand\arraystretch{1.2}
\renewcommand\tabcolsep{2pt}
\begin{tabular}{l|ccc|cccc}
\hline
    & 
    \multicolumn{3}{c|}{PASCAL-S} & \multicolumn{3}{c}{ECSSD} \\
    \multirow{-2}{*}{Methods} & $S_\alpha$ & $E_{\phi}$ & MAE & $S_\alpha$ & $E_{\phi}$ & MAE \\
    \hline
    U2Net~\cite{PR20_U2Net}  & 0.844  & 0.850 & 0.074 & 0.928  & 0.925 & 0.033\\
    ICON~\cite{TPAMI22_ICON}  & 0.861  & 0.893 & 0.064 & 0.929  & 0.954 & 0.032 \\
    EDN~\cite{TIP22_EDN} & 0.865 &  0.902 & 0.062 & 0.927  & 0.951 & 0.032 \\
    MENet~\cite{CVPR23_MENet} & 0.872  & 0.913 & 0.054  & 0.928 & 0.954 & 0.031\\
    \hline
    \textbf{SAM2-UNet} & \textbf{0.894} & \textbf{0.931} & \textbf{0.043} & \textbf{0.950} & \textbf{0.970} & \textbf{0.020} \\  
    \hline
\end{tabular}
\end{table}

\begin{table}[t]
\centering
\caption{Marine animal segmentation performance on MAS3K~\cite{TCSVT21_MAS3K} and RMAS~\cite{JOE23_MASNet} datasets.}
\label{tab:mas}
\renewcommand\arraystretch{1.2}
\renewcommand\tabcolsep{2pt}
\begin{tabular}{l|ccccc|ccccc}
\hline
    & 
    \multicolumn{5}{c|}{MAS3K} & \multicolumn{5}{c}{RMAS}
    \\
    \multirow{-2}{*}{Methods} & $mIoU$ & $S_\alpha$ & $F^{w}_{\beta}$ & $E_{\phi}$ & MAE & $mIoU$ & $S_\alpha$ & $F^{w}_{\beta}$ & $E_{\phi}$ & MAE\\
    \hline
    C2FNet~\cite{IJCAI21_C2FNet} & 0.717 & 0.851 & 0.761 & 0.894 & 0.038 & 0.721 & 0.858 & 0.788 & 0.923& 0.026 \\
    OCENet~\cite{WACV22_OCENet} & 0.667 & 0.824 & 0.703 & 0.868 & 0.052 & 0.680 & 0.836 & 0.752 & 0.900& 0.030\\
    ZoomNet~\cite{CVPR22_ZoomNet} & 0.736 & 0.862 & 0.780 & 0.898 & 0.032 & 0.728 & 0.855 & 0.795 & 0.915 & \textbf{0.022}\\
    MASNet~\cite{JOE23_MASNet} & 0.742 & 0.864 & 0.788 & 0.906 & 0.032 & 0.731 & 0.862 & 0.801 & 0.920 & 0.024\\
    \hline
    \textbf{SAM2-UNet} & \textbf{0.799}	& \textbf{0.903} & \textbf{0.848} & \textbf{0.943} & \textbf{0.021} & \textbf{0.738} & \textbf{0.874} & \textbf{0.810} & \textbf{0.944} & \textbf{0.022}
\\  
    \hline
\end{tabular}
\end{table}

\begin{table}[t]
\centering
\caption{Mirror detection performance on MSD~\cite{ICCV19_MirrorNet} and PMD~\cite{CVPR20_PMD} datasets.}
\label{tab:mirror}
\renewcommand\arraystretch{1.2}
\renewcommand\tabcolsep{2pt}
\begin{tabular}{l|ccc|ccc}
\hline
    & 
    \multicolumn{3}{c|}{MSD} & \multicolumn{3}{c}{PMD}
    \\
    \multirow{-2}{*}{Methods} & $IoU$ & $F$ & MAE & $IoU$ & $F$ & MAE\\
    \hline
    MirrorNet~\cite{ICCV19_MirrorNet} & 0.790 & 0.857 & 0.065 & 0.585 & 0.741 & 0.043 \\
    PMD~\cite{CVPR20_PMD} & 0.815 & 0.892 & 0.047 & 0.660 & 0.794 & 0.032\\
    SANet~\cite{CVPR22_SAMirror} & 0.798 & 0.877& 0.054 & 0.668 & 0.795 & 0.032\\
    HetNet~\cite{AAAI23_HetNet} & 0.828 & 0.906 & 0.043 & 0.690 & 0.814 & 0.029\\
    \hline
    \textbf{SAM2-UNet} & \textbf{0.918} & \textbf{0.957} & \textbf{0.022} & \textbf{0.728} & \textbf{0.826} & \textbf{0.027} \\
    \hline
\end{tabular}
\end{table}

\begin{table}[!htpb]
\centering
\caption{Polyp segmentation performance on Kvasir-SEG~\cite{MMM20_KvasirSEG}, CVC-ClinicDB~\cite{CMIG15_ClinicDB}, CVC-ColonDB~\cite{TMI15_ColonDB}, CVC-300~\cite{Endoscene300}, and ETIS~\cite{ETIS} datasets.}
\label{tab:polyp}
\renewcommand\arraystretch{1.2}
\renewcommand\tabcolsep{1.5pt}
\begin{tabular}{l|cc|cc|cc|cc|cc}
\hline
    & 
    \multicolumn{2}{c|}{Kvasir} & \multicolumn{2}{c|}{ClinicDB} & \multicolumn{2}{c|}{ColonDB} & \multicolumn{2}{c|}{CVC-300} & \multicolumn{2}{c}{ETIS} 
    \\
    \multirow{-2}{*}{Methods} & mDice & mIoU & mDice & mIoU & mDice & mIoU & mDice & mIoU & mDice & mIoU \\
    \hline
    PraNet~\cite{MICCAI20_PraNet} & 0.898 & 0.840 & 0.899 & 0.849 & 0.709 & 0.640 & 0.871 & 0.797 & 0.628 & 0.567
    \\
    SANet~\cite{MICCAI21_SANet} & 0.904 & 0.847 & 0.916 & 0.859 & 0.752 & 0.669  & 0.888 & 0.815 & 0.750 & 0.654 \\
    CaraNet~\cite{SPIE22_CaraNet} & 0.913 & 0.859 & 0.921 & 0.876 & 0.775 & 0.700 & \textbf{0.902} & \textbf{0.836} & 0.740 & 0.660 \\
    CFA-Net~\cite{PR23_CFANet} & 0.915 & 0.861 & \textbf{0.933} & \textbf{0.883} & 0.743 & 0.665 & 0.893 & 0.827 & 0.732 & 0.655 \\
    \hline
    \textbf{SAM2-UNet} & \textbf{0.928} & \textbf{0.879} & 0.907 & 0.856 & \textbf{0.808} & \textbf{0.730} & 0.894 & 0.827 & \textbf{0.796} & \textbf{0.723} \\  
    \hline
\end{tabular}
\end{table}

\subsection{Implementation Details}
Our method is implemented using PyTorch and trained on a single NVIDIA RTX 4090 GPU with 24GB of memory. We use the AdamW optimizer with an initial learning rate of 0.001, applying cosine decay to stabilize training. Two data augmentation strategies are employed: random vertical and horizontal flips. Unless otherwise specified, we use the Hiera-L version of SAM2. All input images are resized to $352 \times 352$, with a batch size of 12. The training epoch is set to 50 for camouflaged object detection and salient object detection, and to 20 for marine animal segmentation, mirror detection, and polyp segmentation. For polyp segmentation, we also adopt a multi-scale training strategy $\{1, 1.25\}$ similar to~\cite{MICCAI20_PraNet}.

\subsection{Comparison with State-of-the-Art Methods}
In this subsection, we first analyze the quantitative results across different benchmarks, followed by visual comparisons in camouflaged object detection and polyp segmentation.

\textbf{Results on Camouflaged Object Detection} are presented in Tables~\ref{tab:cod1} and~\ref{tab:cod2}. SAM2-UNet outperforms all other methods across all four benchmark datasets, achieving the highest scores in every metric. Specifically, in terms of S-measure, SAM2-UNet surpasses FEDER by 1.1\% on the CHAMELEON dataset and by 4.8\% on the CAMO dataset. On the more challenging COD10K and NC4K datasets, which have larger image counts and higher segmentation difficulty, SAM2-UNet still exceeds the performance of FEDER by 3.6\% and 3.9\% in S-measure, respectively.

\textbf{Results on Salient Object Detection} are reported in Tables~\ref{tab:saliency1} and~\ref{tab:saliency2}. SAM2-UNet consistently achieves the top results across all metrics. For S-measure, SAM2-UNet outperforms MENet by 2.9\%, 3.4\%, 1.4\%, 2.2\%, and 2.2\% on the DUTS-TE, DUT-OMRON, HKU-IS, PASCAL-S, and ECSSD datasets, respectively.

\textbf{Results on Marine Animal Segmentation} are detailed in Table~\ref{tab:mas}. Once again, SAM2-UNet achieves the best performance across all metrics on the two benchmark datasets. Specifically, for mIoU, SAM2-UNet outperforms the second-best MASNet by 5.7\% on the MAS3K dataset and by 0.7\% on the RMAS dataset.

\textbf{Results on Mirror Detection} are summarized in Table~\ref{tab:mirror}. SAM2-UNet outshines all other comparison methods in every metric. For instance, SAM2-UNet significantly outperforms HetNet in terms of IoU on the MSD dataset, with a substantial improvement of 9\%. Moreover, on the PMD dataset, SAM2-UNet surpasses HetNet by 3.8\% in IoU.

\textbf{Results on Polyp Segmentation} are shown in Table~\ref{tab:polyp}. SAM2-UNet demonstrates state-of-the-art performance on three out of five datasets. For example, on the Kvasir dataset, SAM2-UNet achieves a mDice score of 92.8\%, surpassing CFA-Net by 1.3\%. Additionally, SAM2-UNet delivers the best performance on ColonDB and ETIS, exceeding CFA-Net by 6.5\% and 6.4\% in mDice. Although our performance is weaker on the ClinicDB and CVC-300 datasets, SAM2-UNet still outperforms CFA-Net by an average of 2.34\% in mDice across all five datasets.

\textbf{Visual Comparison} results are presented in Fig.~\ref{fig:cod} and~\ref{fig:polyp}. In camouflaged object detection, our method demonstrates superior accuracy across various scenes, such as detecting a hidden face (row 1), chameleon (row 2), caterpillar (row 3), and seahorse (row 4). For polyp segmentation, our method effectively reduces false-positive rates (row 1) and false-negative rates (row 2).

\begin{table}[!htpb]
\centering
\caption{Ablation study about different backbones on COD10K~\cite{CVPR20_SINet} and NC4K~\cite{CVPR21_NC4K} datasets.}
\label{tab:backbone}
\renewcommand\arraystretch{1.2}
\renewcommand\tabcolsep{2pt}
\begin{tabular}{l|cccc|cccc}
\hline
     & 
    \multicolumn{4}{c|}{COD10K} & 
    \multicolumn{4}{c}{NC4K}\\
    \multirow{-2}{*}{Backbones}  & $S_\alpha$ & $F_{\beta}$ & $E_{\phi}$  & MAE& $S_\alpha$  & $F_{\beta}$ & $E_{\phi}$ & MAE \\
    \hline
    Hiera-Tiny  & 0.822 & 0.706 & 0.883  & 0.035 & 0.857  & 0.804 & 0.902 & 0.045\\
    Hiera-Small  & 0.839 & 0.729 & 0.900 & 0.031 & 0.869  & 0.822 & 0.913 & 0.040\\
    Hiera-Base+ & 0.853 & 0.749 & 0.910 & 0.027 & 0.879  & 0.833 & 0.920& 0.037\\
    \hline
    \textbf{Hiera-Large} & \textbf{0.880} & \textbf{0.789} & \textbf{0.936}  & \textbf{0.021} & \textbf{0.901}  & \textbf{0.863} & \textbf{0.941} & \textbf{0.029}\\ 
    \hline
\end{tabular}
\end{table}

\subsection{Ablation Study}
To assess the impact of the Hiera backbone size, we conduct ablation experiments, with the results presented in Table~\ref{tab:backbone}. Generally, a larger backbone typically results in better performance. With the smaller Hiera-Base+ backbone, SAM2-UNet still surpasses FEDER and delivers satisfactory results. As the backbone size decreases further, SAM2-UNet also produces results comparable to PFNet and ZoomNet, even with parameter-efficient fine-tuning, demonstrating the high-quality representations provided by the SAM2 pre-trained Hiera backbone.

\section{Conclusion}
In this paper, we propose SAM2-UNet, a simple yet effective U-shaped framework for versatile segmentation across both natural and medical domains. SAM2-UNet is designed for ease of understanding and use, featuring a SAM2 pre-trained Hiera encoder coupled with a classic U-Net decoder. Extensive experiments across eighteen datasets on five benchmarks demonstrate the effectiveness of SAM2-UNet. Our SAM2-UNet can serve as a new baseline for developing future SAM2 variants.

\bibliographystyle{splncs04}
\bibliography{ref}
\end{document}